\documentclass{article}
\usepackage{spconf,amsmath,graphicx}
\usepackage[latin9]{inputenc}
\usepackage[table,xcdraw]{xcolor}
\usepackage{amssymb}
\usepackage{diagbox}

\usepackage{stfloats}

\makeatother

\title{
\vspace{-2cm}
{\hspace{-6.5cm}\footnotesize \textnormal{\textit{Paper accepted to IEEE ICASSP'2018, Calgary, Alberta, Canada, 15-20 April 2018.\\}}
}\vspace{+1.7cm}Yedroudj-Net: An efficient CNN for spatial steganalysis}
\name{Mehdi YEDROUDJ $^{(1)}$\thanks{This work was supported by the Algerian Ministry of Higher Education / Scientific Research, and the University of Montpellier (LIRMM).}, Fr\'ed\'eric COMBY $^{(1)}$, Marc CHAUMONT $^{(1),(2)}$}
\address{
    (1) Montpellier University, LIRMM (UMR5506) / CNRS, FRANCE,\\
    (2) N\^{i}mes University, FRANCE.
}

\begin{document}
\ninept
\maketitle
\begin{abstract}
For about 10 years, detecting the presence of a secret message hidden in an image was performed with an Ensemble Classifier trained with Rich features. In recent years, studies such as Xu {\it et al.} have indicated that well-designed convolutional Neural Networks (CNN) can achieve comparable performance to the two-step machine learning approaches.

In this paper, we propose a CNN that outperforms the state-of-the-art in terms of error probability. The proposition is in the continuity of what has been recently proposed and it is a clever fusion of important bricks used in various papers. Among the essential parts of the CNN, one can cite the use of a pre-processing filter-bank and a Truncation activation function, five convolutional layers with a Batch Normalization associated with a Scale Layer, as well as the use of a sufficiently sized fully connected section. An augmented database has also been used to improve the training of the CNN. 

Our CNN was experimentally evaluated against S-UNIWARD and WOW embedding algorithms and its performances were compared with those of three other methods: an Ensemble Classifier plus a Rich Model, and two other CNN steganalyzers.

\end{abstract}

\begin{keywords}
Steganalysis, Deep Learning, Convolutional Neural Network.
\end{keywords}

%

\section{Introduction}

The first attempt to use Deep Learning methods for steganalysis dates back to 2014 \cite{Tan2014} with auto-encoders. One year later Qian {\it et al.} \cite{Qian_2015_Deep} and Pibre {\it et al.} \cite{Pibre2016} proposed to use Convolutional Neural Networks. In 2016, the first results, close to those of the state-of-the-art, were obtained with an ensemble of CNNs \cite{Xu2016b}. The Xu-Net\footnote{In this paper, we reference {\it Xu-Net} a CNN similar to the one given in \cite{Xu2016a} and not to the ensemble version \cite{Xu2016b}.}\cite{Xu2016a} CNN is used as {\it base learner} of the ensemble of CNNs. Other networks have been proposed in 2017, this time for JPEG steganalysis. In \cite{Zeng2017}, authors proposed a pre-processing inspired by the Rich Models, and the use of a big learning database. The results were close to those of the state-of-the-art. In \cite{Chen2017}, the network is built with a {\it phase-split} inspired by the JPEG compression process. An ensemble of CNNs was required to obtain results that were slightly better than those of the state-of-the-art. In \cite{Xu2017}, a CNN inspired by ResNet \cite{He2016_ResNet} with the {\it shortcut connection} trick and 20 layers also improved the results in term of accuracy.

These results were highly encouraging but regarding the gain obtained in other image processing tasks using Deep Learning methods \cite{Lecun2015}, the steganalysis results were not "10\% better" compared to the classical approaches that use an Ensemble Classifier \cite{Kodovsky2012-EnsembleClassifiers} with a Rich Model \cite{Fridrich2012_Rich, Xia2017} or a Rich Model with a Selection-Channel Awareness \cite{Denemark2014_maxSRM, Denemark2016_SCA}. In 2017, the main trends to improve CNN results are: using an ensemble of CNNs, modifying the topology by mimicking the Rich Models extraction process, or using ResNet. In most of the cases, the design or the experimental effort is very high for a very small improvement of the performance.

By looking back to the {\it good practices} in deep learning as well as the recent studies, we experimentally designed a CNN for spatial steganalysis whose efficiency is naturally better than the state-of-the-art. This is performed without resort to either a design specific to the nature of images (spatial, jpeg, ...) or a CNN ensemble (which is known to improve the results). We focused on the design of the CNN, avoiding the use of tricks known to improve the performances such as transfer learning \cite{Qian2016_Transfer} or virtual augmentation of the database \cite{Ye2017}, etc. Additionally, the proposed network is not sensitive to the initialization of hyperparameters and thus easily converges, which will be later discussed in Section~\ref{batchNormalisation}.
We named this network the "Yedroudj-Net" CNN, and will compare it with Xu-Net \cite{Xu2016a}, Ye-Net \cite{Ye2017} and also with the Ensemble Classifier \cite{Kodovsky2012-EnsembleClassifiers} fed with the Spatial-Rich-Models \cite{Fridrich2012_Rich} for spatial steganalysis.

\section{Yedroudj-Net}

\begin{figure*}[t]
\centering
\includegraphics[width=18cm,height=4cm]{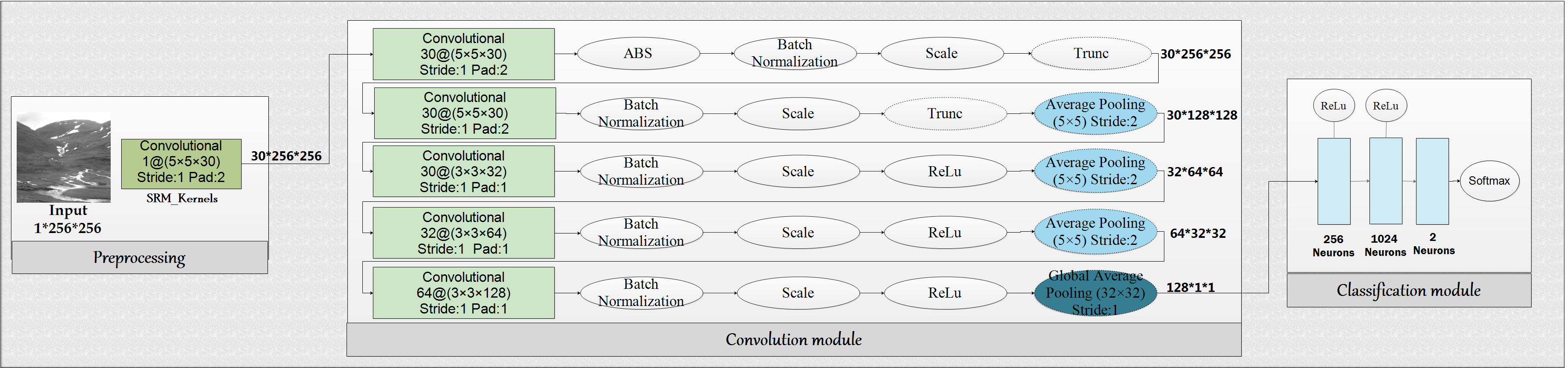} 
\caption{Yedroudj-Net CNN architecture.}
\label{fig:yedroudj-net}
\end{figure*}

Fig. \ref{fig:yedroudj-net} illustrates the overall architecture of our CNN. The network is composed of a {\it pre-processing block}, five {\it convolutional blocks}, and a {\it fully connected block} made of three fully connected layers followed by a {\it softmax}. The network produces a probability distribution over the two class labels.

The {\it pre-processing block} filters the input cover/stego image with a predefined high-pass filter in order to extract the noise component residuals. The pre-processed image then feeds the network. Previous studies \cite{Qian_2015_Deep, Pibre2016} observed that without this preliminary high-pass filter the CNN converges more slowly. This pre-processing largely suppresses the image content, narrows the dynamic range, and thus increases the signal-to-noise ratio between the weak stego signal (if present) and the image signal. As a result, the CNN can learn on a more compact and {\it robust} signal. 

Inspired by the benefit of {\it diversity} \cite{Fridrich2012_Rich}, and similarly to \cite{Ye2017}, we use the 30-basic high-pass filters from SRM \cite{Fridrich2012_Rich}, instead of using only one filter such as \cite{Qian_2015_Deep, Pibre2016,Xu2016a}, in order to pre-process the input image. 
Note that the filters kernel values of the {\it preprocessing block}, i.e. the weights, are not optimized/learned during the training. This pre-processing has been integrated into a lazy fashion, directly into the CNN, such that the size of all kernels (weighting matrix) are set to 5$\times$5. Their central part is initialized with the weights of the SRM kernels and the remaining elements are padded to zero. No normalization of the kernels' values is performed.

The rest of our CNN can be divided into a convolutional module, dedicated to features representation, that transforms the input image into a {\it feature vector}, and a classification module, consisting of three fully-connected layers and a softmax layer, which produces the classification decision (cover or stego).

Similarly to Xu-Net, the convolutional module has five blocks marked as 'Block 1' through 'Block 5' to extract effective features for cover and stego images discrimination; see Fig. \ref{fig:yedroudj-net}. Each block is made of the following steps:
\begin{enumerate}
\item a {\it Convolution Layer}. Similar to Xu-Net \cite{Xu2016a}, we set the size of the convolutional kernels to 5$\times$5 for Blocks 1 and 2, but we reduced it to 3$\times$3 for the Blocks 3 through 5. For all the convolution layers and similarly to Res-Net \cite{He2016_ResNet} and Xu-Net \cite{Xu2016a}, no biases are used. Biases terms are set to false on the convolution layer and moved to the Scale Layer.

\item an {\it Absolute Value} activation (ABS) layer. This ABS layer is only used in Block 1  similarly to Xu-Net. It forces the statistical modeling to consider the sign symmetry of the noise residuals. The relevance of this layer was observed in Xu-Net \cite{Xu2016a}.

\item a {\it Batch Normalization (BN)}.\label{batchNormalisation} The BN 
normalizes the distribution of each feature to a zero-mean and a unit-variance, and eventually, scales and translates the distribution. The benefit of using a BN layer is that it desensitizes the training to the parameters initialization \cite{batch}, allows the use of a larger learning rate which speeds up the learning, and improves the detection accuracy \cite{Chen2017}. Note that similarly to ResNet \cite{He2016_ResNet}, and in contrast to Xu-Net, we provide a BN layer accompanied by a scale layer. The latter attempts to learn the scaling and translation parameters more efficiently. Those two parameters can be well learned by the independent {\it Scale Layer}. Similarly to ResNet, we observe a very slight increase in the network's accuracy.

\item a non-linear {\it Activation layer}. For the Blocks 1 and 2, a {\it Truncation function} is used to limit the range of data values and prevent the deeper layers from modeling large values. Indeed, these values are sparse and not statistically significant. The formula of the truncation function ({\it Trunc}) is given in Eq. \ref{Trunc}, and is parameterized by $T\in \mathbb{N}$, a threshold:
\begin{equation}
\label{Trunc}
Trunc(x)= 
\begin{cases}
    \textit{-T},             & x < -T,\\
    \textit{x},    & -T\leq x\leq T,\\
    \textit{T},              & x > T.
\end{cases}
\end{equation}   

This {\bf outlier} suppression process, proposed in \cite{Ye2017}, can also be seen as the use of a {\it robustness function}. For the Blocks, 3 through 5, the classical {\it Rectified Linear Unit} (ReLU) is used because it yields good performances and its gradient computation is fast. 

\item An {\it Average pooling}. This average pooling layer is exclusively used in Blocks 2 through 5. This allows to down-sample the feature maps, and thus reduces the dimensionality. For the last block, a {\it global} average pooling is performed to generate a one by one element for each corresponding feature map, thereby preventing the statistical modeling from grasping the location information of embedded pixels from the training data \cite{Average}. There is no pooling in the first block to avoid information loss at the beginning of the network.
\end{enumerate}

The features extracted from the convolutional module feed the classification module which consists of three fully connected layers. The number of neurons in the first and second layers is 256 and 1024 respectively, and the last fully connected layer has only two neurons corresponding to the number of classes of the network's output. At the end of this module, a softmax activation function is used to produce a distribution over the two class labels.

\begin{figure*}[!htb]
\centering
\includegraphics[width=18cm,height=5cm]{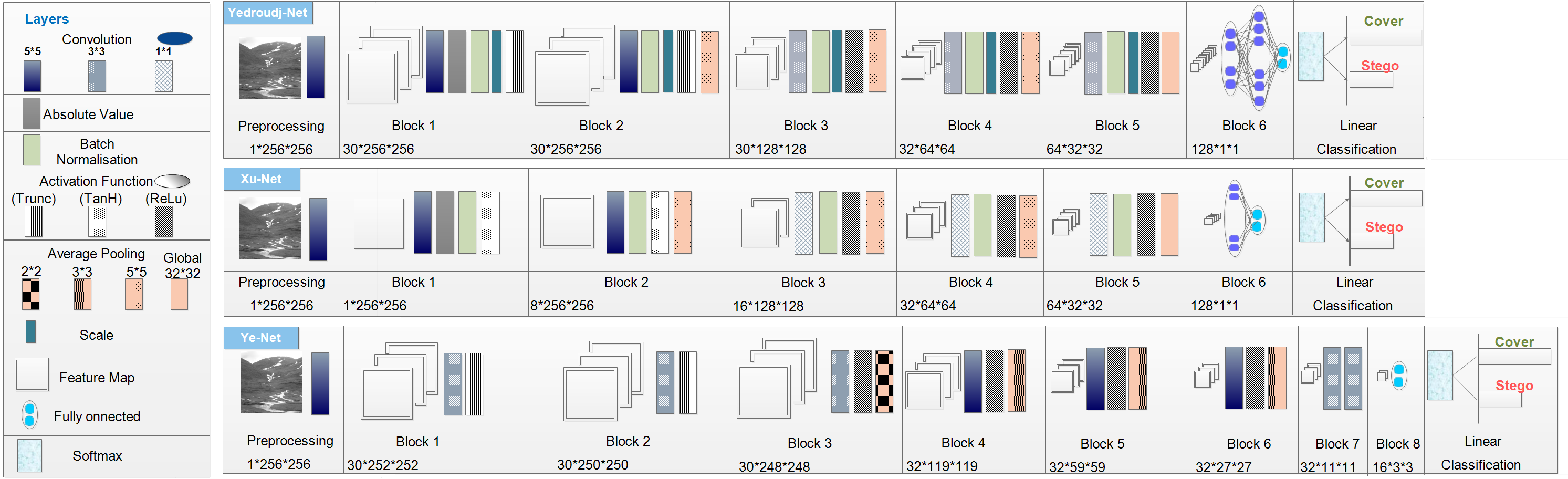} 
\caption{Comparison of Yedroudj-Net, Xu-Net, and Ye-Net architectures.}
\label{fig:nets}
\end{figure*}
\section{Experiments}
\label{sec:experiments}


\subsection{Dataset and software platform}

We use S-UNIWARD \cite{Holub2014}, and WOW \cite{Holub2012_WOW}, two well-known content-adaptive methods for the embedding in the spatial domain and their  Matlab implementations (online codes\footnote{http://dde.binghamton.edu/download/}) with the simulator for the embedding and a random key for each embedding. We thus avoid any wrong use of the C++ codes, i.e. a fixed and unique embedding key, as reported in \cite{Pibre2016}.

Our steganalysis CNN, {\it Yedroudj-Net}, is compared with the state-of-the-art approaches: {\it Xu-Net} CNN \cite{Xu2016a}, {\it Ye-Net} CNN \cite{Ye2017}, and with {\it SRM + EC} which stands for the hand-crafted feature set Spatial-Rich-Model \cite{Fridrich2012_Rich} and the Ensemble Classifier \cite{Kodovsky2012-EnsembleClassifiers}. For a fair comparison, all the involved steganalysis methods are tested on the same subsampled images from the BOSSBase database v.1.01 \cite{Bas2011-BOSS}. All CNNs experiments were performed with the publicly available {\it Caffe} toolbox \cite{caffe_jia} with necessary modifications, plus digits V5. All tests were run on an NVidia Titan X GPU card.

\subsection{Training, Validation, Test}
\label{Training_section}

Due to our GPU computing platform and time limitation, we conduct all the experiments on images of 256$\times$256 pixels, similarly to \cite{Ye2017}. To this end, we resampled all the 512$\times$512 images to 256$\times$256 images, using the {\it imresize()} Matlab function with the default parameters.
Then, our 256$\times$256 BOSSBase is split into two sets,  50\% (resp. the other 50\%) of the cover/stego pairs is assigned to the training (resp. testing) set. 4000 out of the 5000 training set pairs are randomly selected for training, the remaining 1000 pairs are set aside for validation. The testing set is left untouched during the training stage. 

During the CNNs training, we fixed a maximum of 900 epochs. Nevertheless, most of the time, we manually stopped the training when an over-fitting phenomenon appeared (usually before the epoch 200 for WOW and 300 for S-UNIWARD), i.e. when the {\it Loss} continues to decrease on the training set but starts to increase on the validation set. In practice, observing the {\it Loss} curve computed on the validation test leads us to keep two versions of the CNN: the CNN's models with minimum Loss (resp. maximum) on the validation set over the previous five epochs. Those two CNN's models are evaluated on the testing set, and we report the average error probability of detection for these two CNN's models.

For {\it SRM + EC} we use the SRM feature set of dimension=34 671 \cite{Fridrich2012_Rich}, and the Ensemble Classifier \cite{Kodovsky2012-EnsembleClassifiers}. We report the minimum error probability under equals prior, averaged over 10 tests.


\subsection{Hyper-parameters} 

We apply a mini-batch stochastic gradient descent (SGD) to train our CNN. The momentum is fixed to 0.95 and the weight decay to 0.0001. No dropout is used. 
The batch size in the training procedure is set to 16, due to GPU memory limitation (8 cover/stego pairs). All layers are initialized using {\bf Xavier method}: the weights follow a Gaussian distribution and are chosen so that the variance for both input and output among each layer remains the same \cite{Glorot2010}. During the training, we use the {\it step policy} of Caffe to adjust the learning rate (initialized to 0.01). With this {\it policy}, each 10\% of the total number of epochs, our learning rate is decreased by a factor gamma equal to 0.1. The threshold $T$, for the Truncations functions (see Equ.~\ref{Trunc}) is set to 3 for the first layer and 2 for the second layer, and the 30-basic high-pass SRM filters are not normalized. Note that the source codes and the materials files are available at http://www.lirmm.fr/ $\tilde{ }$ chaumont/DemoAndSources.html.

\subsection{Difference between the 3 CNNs}
In this section we will briefly discuss the differences between our CNN {\it Yedroudj-Net}, the {\it Xu-Net} CNN and {\it Ye-Net} CNN, the state-of-the-art CNNs for the spatial steganalysis. In our comparisons, {\it Xu-Net} is a CNN similar to the one given in \cite{Xu2016a} that takes images of size of 256$\times$256 instead of 512$\times$512. We thus suppressed the {\it average pooling} from the first Block, which is a favorable measure since it avoids an early down-sampling. We also set a ReLU activation function among the Fully connected layers. Fig.\ref{fig:nets} shows the overall architectures of all CNNs. We summarize below the major similarities and differences between the CNNs:
\begin{itemize}
\item Both Yedroudj-Net and Xu-Net use 5 convolution layers. Yedroudj-Net has nevertheless two times more features (256) at the input of the fully connected section. Ye-Net has more convolution layers.
\item Both Yedroudj-Net and Xu-Net use a {\it Batch Normalization} layer; the Ye-Net does not.
\item Both Yedroudj-Net and Xu-Net use the {\it Absolute Value} layer; the Ye-Net does not.
\item Both Yedroudj-Net and Ye-Net use a 30 filter bank for pre-processing; the Xu-Net does not
\item Both Yedroudj-Net and Ye-Net Net use a Truncation activation function in Block 1 and 2 (We have found "Experimentally" that using Truncation activation function only in the Blocks 1 and 2 is the best choice in term of detection accuracy, those experiments are not reported here); the Xu-Net does not.
\item Yedroudj-Net has three (resp. Xu-Net two, and Ye-Net one) fully connected layer.
\end{itemize}
 
\subsection{Results without using any tricks}


\subsubsection{General performance comparisons}

In Table \ref{tab:comparison}, we report the error probability obtained when steganalyzing WOW and S-UNIWARD embedding algorithms at 0.2 bpp and 0.4 bpp. The steganalysis methods are Yedroudj-Net, Xu-Net, Ye-Net, and SRM+EC \cite{Kodovsky2012-EnsembleClassifiers, Fridrich2012_Rich}.

\begin{table}[htb]
\centering
\renewcommand{\arraystretch}{1.6}
\caption{
Steganalysis error probability comparison of Yedroudj-Net, Xu-Net, Ye-Net, and SRM+EC for two embedding algorithms WOW and S-UNIWARD at 0.2 bpp and 0.4 bpp.}
\label{tab:comparison}
\renewcommand{\arraystretch}{1.1}
\scalebox{0.8}{
\begin{tabular} {l|c|c|c|c|} 
\cline{2-5} 
                                                                                 & \multicolumn{4}{c|}{\cellcolor[HTML]{B4AAAA}{ \textit{\textbf{BOSS 256$\times$256}}}}                           \\ \cline{2-5} 
                                                                                   & \multicolumn{2}{c|}{\cellcolor[HTML]{B4ACAC}\textbf{WOW} \cite{Holub2012_WOW}} & \multicolumn{2}{c|}{\cellcolor[HTML]{B4ACAC}\textbf{S-UNIWARD} \cite{Holub2014}} \\ \hline
\multicolumn{1}{|l|}{{\cellcolor[HTML]{B4ACAC}\backslashbox {\textbf{Steganalysis} }{\textbf{Payload}}}}                                 
& {\cellcolor[HTML]{C0C0C0} 0.2 bpp} & {\cellcolor[HTML]{C0C0C0} 0.4 bpp} & {\cellcolor[HTML]{C0C0C0} 0.2 bpp} & {\cellcolor[HTML]{C0C0C0} 0.4 bpp} \\ \hline
\multicolumn{1}{|l|}{\cellcolor[HTML]{C0C0C0}{SRM+EC \cite{Kodovsky2012-EnsembleClassifiers, Fridrich2012_Rich}}}       & 36.5 \%                       & 25.5 \%                       & {\bf 36.6} \%                           & 24.7 \%                          \\ \hline
\multicolumn{1}{|l|}{\cellcolor[HTML]{C0C0C0}{Yedroudj-Net }} & {\bf 27.8} \%                      & {\bf 14.1} \%                       & {\bf 36.7} \%                           & {\bf 22.8} \%                          \\ \hline
\multicolumn{1}{|l|}{\cellcolor[HTML]{C0C0C0}{Xu-Net \cite{Xu2016a}}}       & 32.4 \%                       & 20.7 \%                       & 39.1 \%                           & 27.2 \%                          \\ \hline
\multicolumn{1}{|l|}{\cellcolor[HTML]{C0C0C0}{Ye-Net \cite{Ye2017}}}       & 33.1 \%                      & 23.2 \%                       & 40.0 \%                           & 31.2 \%                          \\ \hline
\end{tabular}}
\vspace{-0,5cm}
\end{table}
For WOW algorithm, Yedroudj-Net has an error probability 8\% lower (resp. 11\%) at 0.2 bpp (resp. 0.4 bpp) compared to SRM+EC. The results are also favorable for S-UNIWARD steganalysis with an equal error probability at 0.2 bpp and 2\% lower at 0.4 bpp.

Compared to the other CNN algorithms, our proposed CNN achieves far superior results. Yedroudj-Net is 2\% to 6\% better compared to Xu-Net for the two embedding algorithms and the two payloads. The results are even better when compared to Ye-Net, where Yedroudj-Net is 3\% to 9\% better. Let us note that the two other CNNs are not always superior when compared to the SRM+EC. To beat SRM+EC, those approaches require using an ensemble of CNN, as proposed in \cite{Xu2016b}, or increasing the learning database, as proposed in \cite{Zeng2017}, and showed in section below. 

Note that extreme caution must be taken for the initialization of the learning rate of the Ye-Net and the management of its evolution through the epochs. Indeed, a bad initialization prevents the network from converging. In Yedroudj-Net and Xu-Net, the use of the Batch Normalization ensures less sensitivity to such a parameter setting.

To conclude on these general comparisons, in a classical clairvoyant scenario without any channel-awareness, and without using an ensemble, a larger database, a virtual augmentation of the database, or a transfer learning, Yedroudj-Net has a clear advantage over all the state-of-the-art methods.

\subsection{Results with a Base augmentation}
Many tricks exist for improving the results of CNN but the {\it base augmentation} seems to be a very important measure to apply in order to better exploit the capacity of Deep Learning approaches.

\begin{table}[htb]
\centering
\caption{Base Augmentation influence: error probability comparison of  Yedroudj, Xu and Ye nets on WOW at 0.2 bpp with a learning base augmented with BOWS2, and Virtually Augmented.}
\label{Base Augmentation}
\renewcommand{\arraystretch}{1.1}
\scalebox{0.9}{
\begin{tabular}{l|l|l|l|}
\cline{2-4}
                                                     & \cellcolor[HTML]{C0C0C0}BOSS & \cellcolor[HTML]{C0C0C0}BOSS+BOWS2 & \cellcolor[HTML]{C0C0C0}BOSS+BOWS2+VA \\ \hline
\multicolumn{1}{|l|}{\cellcolor[HTML]{C0C0C0}Yedroudj-Net}   & {\bf 27.8} \%                        & {\bf 23.7} \%                             & {\bf 20.8} \%                             \\ \hline
\multicolumn{1}{|l|}{\cellcolor[HTML]{C0C0C0}Ye-Net} & 33.1 \%                        & 26.1 \%                              & 22.2 \%                                      \\ \hline
\multicolumn{1}{|l|}{\cellcolor[HTML]{C0C0C0}Xu-Net}     & 32.4 \%                        & 30.3 \%                                 & 30.5 \%                                      \\ \hline
\end{tabular}}
\end{table}



In machine learning, and this is also true for CNNs, it is important to use a training base large enough to ensure a good generalization but also to avoid over-training. Some authors are prone to use big databases \cite{Qian_2015_Deep, Zeng2017,Ye2017} in order to reach the state-of-the-art results. In the above experiment, we attempt to investigate the improvement brought by increasing the learning database size without modifying the testing set. It means that the learning set does not only contain images of the same kind as in the test set: e.g. the settings of cameras, the scenes of the learning set, can all be different from those of the testing set. We show the effects of increasing the image database on the error probability in Table\ref{Base Augmentation}. To increase the size of our training set, two scenarios have been tested inspired by \cite{Ye2017}. 

In the first scenario, noted BOSS+BOWS2, we embedded the payload in the subsampled BOSSBase database v.1.01 \cite{Bas2011-BOSS}. We split this base into two sets: 50\% of the cover/stego pairs to the training set, the rest to the testing set. Then, 10 000 additional pairs of cover/stego pair (obtained by subsampling BOWS2Base \cite{BOWS2008}) were added to the training set. The learning database now contains 15 000 pairs of cover/stego images minus 1000 pairs from BOSS, set aside for validation.

In the second scenario, noted BOSS+BOWS2+VA, the database is virtually augmented by performing the label-preserving flips and rotations on the BOSS+BOWS2 training set. The size of the BOSS+BOWS2 training set is thus increased by a factor of 8, which virtually gives a final learning database made of 112 000 pairs of cover/stego images plus 1000 pairs from BOSS used for validation.

Table \ref{Base Augmentation} shows the performance comparisons in terms of detection error probability for Yedroudj-Net, Xu-Net \cite{Xu2016a}, Ye-Net \cite{Ye2017}, against the embedding algorithm WOW \cite{Holub2012_WOW} at payload 0.2 bpp. For all algorithms, better performances are achieved using BOSS+BOWS2 compared to using only  BOSSBase. The Yedroudj-Net obtains the best results and decreases its detection error probability by 4\%. Ye-Net and Xu-Net respectively decrease their detection error probability by 7\% and 2\%. At this point, it was not clear if the improvement was only due to a lack of data or also because the additional images came from the same cameras. We have nevertheless conducted additional experiments, reported in the paper \cite{Yedroudj2018_DatabaseAugmentation}, and it seems that in order to improve the performance, one must increase the database with images coming from the same sources and with a development process respecting the pixels resolutions and ratios.


When virtually augmenting the entire BOSS+BOWS2 learning set (i.e. BOSS+BOWS2+VA) thanks to the 8 combinations of rotations and flips that do not introduce interpolation, the performances are again increased. The Yedroudj-Net keeps the best results and decreases its detection error probability by 7\% (Ye-Net decreases it by 11\%, and Xu-Net by 2\%) compared to the case of only using BOSSBase for the training. Comparing to RM+EC \cite{Kodovsky2012-EnsembleClassifiers, Fridrich2012_Rich}, whose error probability is 36.5\% with a learning on the BOSSBase, the Yedroudj-Net obtain an error probability of 20.8\% which give an improvement of 16\%. The Ye-Net obtains an improvement of 14\% and the Xu-Net an improvement of 6\%.

These tests reveal how important it is to have a large database when using CNN of 5-7 blocks. The number of parameters (without taking into account the BN and/or scale) goes approximately from 50 thousand (Xu-Net) to 500 thousand (Yedroudj-Net). Such a huge number of unknown requires bearing enough samples. The experiments show that the CNNs still do not have enough learning samples. For a steganalysis of BOSSBase with CNNs of 5-7 blocks, even 112 000 pairs of images (BOSS+BOWS2 virtually augmented) is not enough. Consequently, using a bigger base allows our CNN to achieve better performances even if the convergence time increases. 

Using a GPU card of the previous generation (Nvidia TitanX) on an Intel Core i7-5930K CPU 3.50GHz$\times$12 with 32G of RAM, it takes less than one day for learning Yedroudj-Net CNN on BOSSBase, three days on BOSS+BOWS2, and more than seven days on BOSS+BOWS2+VA.


\section{Conclusion}

This article presents the evaluation of the Yedroudj-Net CNN, designed for spatial steganalysis. This CNN gathers some recent design propositions in order to build a simple approach beating the state-of-the-art approaches in a classical clairvoyant scenario without knowledge of the selection channel. 

The key to the steganalysis performance improvement is the combination of the following elements: a bank of filters for the pre-processing step, a Truncation activation function, and a Batch Normalization associated with a Scale Layer.

An additional experiment dealing with the problem of the learning base size showed that by adding BOWS2 and virtually augmenting the learning database, the results become extremely satisfactory. An experiment on WOW at 0.2 bpp led to an error probability decrease of 16\% compared to the RM+EC.


\bibliographystyle{IEEEbib}
{
\footnotesize   
\bibliography{biblio1}

\begin{thebibliography}{10}

\bibitem{Tan2014}
S.~Tan and B.~Li,
\newblock ``Stacked convolutional auto-encoders for steganalysis of digital
  images,''
\newblock in {\em Proceedings of Signal and Information Processing Association
  Annual Summit and Conference, APSIPA'2014}, Siem Reap, Cambodia, Dec. 2014,
  pp. 1--4.

\bibitem{Qian_2015_Deep}
Yinlong Qian, Jing Dong, Wei Wang, and Tieniu Tan,
\newblock ``{Deep Learning for Steganalysis via Convolutional Neural
  Networks},''
\newblock in {\em Proceedings of Media Watermarking, Security, and Forensics
  2015, MWSF'2015, Part of IS\&T/SPIE Annual Symposium on Electronic Imaging,
  SPIE'2015}, San Francisco, California, USA, Feb. 2015, vol. 9409, pp.
  94090J--94090J--10.

\bibitem{Pibre2016}
L.~Pibre, J.~Pasquet, D.~Ienco, and M.~Chaumont,
\newblock ``Deep learning is a good steganalysis tool when embedding key is
  reused for different images, even if there is a cover source-mismatch,''
\newblock in {\em Proceedings of Media Watermarking, Security, and Forensics,
  MWSF'2016, Part of I\&ST International Symposium on Electronic Imaging,
  EI'2016}, San Francisco, California, USA, Feb. 2016, pp. 1--11.

\bibitem{Xu2016b}
Guanshuo Xu, Han-Zhou Wu, and Yun~Q. Shi,
\newblock ``{Ensemble of CNNs for Steganalysis: An Empirical Study},''
\newblock in {\em Proceedings of the 4th ACM Workshop on Information Hiding and
  Multimedia Security}, Vigo, Galicia, Spain, June 2016, IH\&MMSec'16, pp.
  103--107.

\bibitem{Xu2016a}
G.~Xu, H.~Z. Wu, and Y.~Q. Shi,
\newblock ``{Structural Design of Convolutional Neural Networks for
  Steganalysis},''
\newblock {\em IEEE Signal Processing Letters}, vol. 23, no. 5, pp. 708--712,
  May 2016.

\bibitem{Zeng2017}
Jishen Zeng, Shunquan Tan, Bin Li, and Jiwu Huang,
\newblock ``{Pre-training via fitting deep neural network to rich-model
  features extraction procedure and its effect on deep learning for
  steganalysis},''
\newblock in {\em Proceedings of Media Watermarking, Security, and Forensics
  2017, MWSF'2017, Part of IS\&T Symposium on Electronic Imaging, EI'2017},
  Burlingame, California, USA, Jan. 2017, p.~6.

\bibitem{Chen2017}
Mo~Chen, Vahid Sedighi, Mehdi Boroumand, and Jessica Fridrich,
\newblock ``{JPEG-Phase-Aware Convolutional Neural Network for Steganalysis of
  JPEG Images},''
\newblock in {\em Proceedings of the 5th ACM Workshop on Information Hiding and
  Multimedia Security}, Drexel University in Philadelphia, PA, June 2017,
  IH\&MMSec'17, pp. 75--84.

\bibitem{Xu2017}
Guanshuo Xu,
\newblock ``{Deep Convolutional Neural Network to Detect J-UNIWARD},''
\newblock in {\em Proceedings of the 5th ACM Workshop on Information Hiding and
  Multimedia Security}, Drexel University in Philadelphia, PA, June 2017,
  IH\&MMSec'17, pp. 67--73.

\bibitem{He2016_ResNet}
Kaiming He, Xiangyu Zhang, Shaoqing Ren, and Jian Sun,
\newblock ``Deep residual learning for image recognition,''
\newblock in {\em Proceedings of IEEE Conference on Computer Vision and Pattern
  Recognition, CVPR'2016}, Las Vegas, Nevada, June 2016, pp. 770--778.

\bibitem{Lecun2015}
Yann LeCun, Yoshua Bengio, and Geoffrey Hinton,
\newblock ``Deep learning,''
\newblock {\em Nature}, vol. 521, no. 7553, pp. 436--444, May 2015.

\bibitem{Kodovsky2012-EnsembleClassifiers}
J.~Kodovsk{\'y}, J.~Fridrich, and V.~Holub,
\newblock ``{Ensemble Classifiers for Steganalysis of Digital Media},''
\newblock {\em IEEE Transactions on Information Forensics and Security, TIFS},
  vol. 7, no. 2, pp. 432--444, 2012.

\bibitem{Fridrich2012_Rich}
J.~Fridrich and J.~Kodovsk\'y,
\newblock ``{Rich Models for Steganalysis of Digital Images},''
\newblock {\em IEEE Transactions on Information Forensics and Security, TIFS},
  vol. 7, no. 3, pp. 868--882, June 2012.

\bibitem{Xia2017}
Chao Xia, Qingxiao Guan, Xianfeng Zhao, Zhoujun Xu, and Yi~Ma,
\newblock ``{Improving GFR Steganalysis Features by Using Gabor Symmetry and
  Weighted Histograms},''
\newblock in {\em Proceedings of the 5th ACM Workshop on Information Hiding and
  Multimedia Security}, Drexel University in Philadelphia, PA, June 2017,
  IH\&MMSec'17, p.~11.

\bibitem{Denemark2014_maxSRM}
T.~Denemark, V.~Sedighi, V.~Holub, R.~Cogranne, and J.~Fridrich,
\newblock ``Selection-channel-aware rich model for steganalysis of digital
  images,''
\newblock in {\em Proceedings of IEEE International Workshop on Information
  Forensics and Security, WIFS'2014}, Atlanta, Georgia, USA, Dec. 2014, pp.
  48--53.

\bibitem{Denemark2016_SCA}
T.~Denemark, M.~Boroumand, and J.~Fridrich,
\newblock ``Steganalysis features for content-adaptive jpeg steganography,''
\newblock {\em IEEE Transactions on Information Forensics and Security}, vol.
  11, no. 8, pp. 1736--1746, Aug. 2016.

\bibitem{Qian2016_Transfer}
Y.~Qian, J.~Dong, W.~Wang, and T.~Tan,
\newblock ``Learning and transferring representations for image steganalysis
  using convolutional neural network,''
\newblock in {\em Proceedings of IEEE International Conference on Image
  Processing, ICIP'2016}, Phoenix, Arizona, Sept. 2016, pp. 2752--2756.

\bibitem{Ye2017}
Jian Ye, Jiangqun Ni, and Y.~Yi,
\newblock ``Deep learning hierarchical representations for image
  steganalysis,''
\newblock {\em IEEE Transactions on Information Forensics and Security, TIFS},
  vol. 12, no. 11, pp. 2545--2557, Nov. 2017.

\bibitem{batch}
Sergey Ioffe and Christian Szegedy,
\newblock ``Batch normalization: Accelerating deep network training by reducing
  internal covariate shift,''
\newblock in {\em Proceedings of the 32nd International Conference on Machine
  Learning, {ICML} 2015, Lille, France, 6-11 July 2015}, 2015, pp. 448--456.

\bibitem{Average}
Min Lin, Qiang Chen, and Shuicheng Yan,
\newblock ``Network in network,''
\newblock in {\em International Conference on Learning Representations, ICLR
  2014}, Banff, Canada, Apr. 2014, p.~10.

\bibitem{Holub2014}
V.~Holub, J.~Fridrich, and T.~Denemark,
\newblock ``{Universal Distortion Function for Steganography in an Arbitrary
  Domain},''
\newblock {\em EURASIP Journal on Information Security, JIS}, vol. 2014, no. 1,
  2014.

\bibitem{Holub2012_WOW}
V.~Holub and J.~Fridrich,
\newblock ``{Designing Steganographic Distortion Using Directional Filters},''
\newblock in {\em Proceedings of the IEEE International Workshop on Information
  Forensics and Security, WIFS'2012}, Tenerife, Spain, Dec. 2012, pp. 234--239.

\bibitem{Bas2011-BOSS}
P.~Bas, T.~Filler, and T.~Pevn{\'y},
\newblock ``{'Break Our Steganographic System': The Ins and Outs of Organizing
  BOSS},''
\newblock in {\em Proceedings of the 13th International Conference on
  Information Hiding, IH'2011}, Prague, Czech Republic, May 2011, vol. 6958 of
  {\em Lecture Notes in Computer Science}, pp. 59--70, Springer.

\bibitem{caffe_jia}
Yangqing Jia, Evan Shelhamer, Jeff Donahue, Sergey Karayev, Jonathan Long, Ross
  Girshick, Sergio Guadarrama, and Trevor Darrell,
\newblock ``Caffe: Convolutional architecture for fast feature embedding,''
\newblock in {\em Proceedings of the 22nd ACM international conference on
  Multimedia}, Orlando, Florida, USA, Nov. 2014, ACM, pp. 675--678.

\bibitem{Glorot2010}
Xavier Glorot and Yoshua Bengio,
\newblock ``Understanding the difficulty of training deep feedforward neural
  networks,''
\newblock in {\em Proceedings of the Thirteenth International Conference on
  Artificial Intelligence and Statistics, AISTATS'2010}, Chia Laguna Resort,
  Sardinia, Italy, May 2010, vol.~9 of {\em Proceedings of Machine Learning
  Research}, pp. 249--256.

\bibitem{BOWS2008}
P.~Bas and T.~Furon,
\newblock ``{BOWS-2 Contest (Break Our Watermarking System)},'' 2008,
\newblock Organized between the 17th of July 2007 and the 17th of April 2008.
  http://bows2.ec-lille.fr/.

\bibitem{Yedroudj2018_DatabaseAugmentation}
Mehdi Yedroudj, Marc Chaumont, and Fr\'ed\'eric Comby,
\newblock ``{How to augment a small learning set for improving the performances
  of a CNN-based steganalyzer?},''
\newblock in {\em Proceedings of Media Watermarking, Security, and Forensics,
  MWSF'2018, Part of IS\&T International Symposium on Electronic Imaging,
  EI'2018}, Burlingame, California, USA, 28 Jan - 2 Feb 2018.

\end{thebibliography}
}

\end{document}